\documentclass[final,5p,times,twocolumn]{elsarticle}

\usepackage[utf8]{inputenc}        
\usepackage[T1]{fontenc}           
\usepackage{lmodern}               

\usepackage{amsmath}        
\usepackage{amssymb}        
\usepackage{amsthm}         
\usepackage{amsfonts}       
\usepackage{mathrsfs}       
\usepackage{bm}             

\usepackage{booktabs}       
\usepackage{array}          
\usepackage{algorithm}      
\usepackage{algpseudocode} 

\usepackage{graphicx}       
\usepackage{multirow}       
\usepackage{tabularx}       
\usepackage{colortbl}       
\usepackage{xcolor}         
\usepackage{diagbox}        
\usepackage{pifont}         
\usepackage{makecell}       
\usepackage{wrapfig}        
\usepackage{threeparttable} 
\usepackage{nicefrac}       
\usepackage{microtype}      
\usepackage{placeins}       
\usepackage{float}          
\usepackage{stfloats}       
\usepackage{svg}            
\usepackage{caption}        
\usepackage{subcaption}     
\usepackage{booktabs}    
\usepackage{multirow}    
\usepackage{siunitx}     

\begin{document}

\begin{frontmatter}

\title{DIFFUMA: High-Fidelity Spatio-Temporal Video Prediction via Dual-Path Mamba and Diffusion Enhancement}

\author[1]{Xinyu Xie}
\ead{x13817775216@163.com}
\author[1]{Weifeng Cao\corref{corauthor}}
\ead{caoweifeng@zzuli.edu.cn}
\author[1]{Jun Shi}
\ead{shijunzz@gmail.com}
\author[2]{Yangyang Hu}
\ead{x18964114636@163.com}
\author[1]{Hui Liang}
\ead{hliang@zzuli.edu.cn}
\author[1]{Wanyong Liang}
\ead{24676248@qq.com}
\author[1]{Xiaoliang Qian}
\ead{qxl_sunshine@163.com}

\cortext[corauthor]{Corresponding author}

\affiliation[1]{organization={College of Electric and Information Engineering, Zhengzhou University of Light Industry},
            city={Zhengzhou},
            postcode={450002},
            country={PR China}}
            
\affiliation[2]{organization={Guangli Ruihong Electronic Technology Co., Ltd},
            city={Zhengzhou},
            postcode={450046},
            country={PR China}}

\begin{abstract}
Spatio-temporal video prediction plays a pivotal role in critical domains, ranging from weather forecasting to industrial automation. However, in high-precision industrial scenarios such as semiconductor manufacturing, the absence of specialized benchmark datasets severely hampers research on modeling and predicting complex processes. To address this challenge, we make a twofold contribution. \textbf{First, we construct and release the Chip Dicing Lane Dataset (CHDL)}, the first public temporal image dataset dedicated to the semiconductor wafer dicing process. Captured via an industrial-grade vision system, CHDL provides a much-needed and challenging benchmark for high-fidelity process modeling, defect detection, and digital twin development. \textbf{Second, we propose DIFFUMA}, an innovative dual-path prediction architecture specifically designed for such fine-grained dynamics. The model captures global long-range temporal context through a parallel Mamba module, while simultaneously leveraging a diffusion module, guided by temporal features, to restore and enhance fine-grained spatial details, effectively combating feature degradation. Experiments demonstrate that on our CHDL benchmark, DIFFUMA significantly outperforms existing methods, reducing the Mean Squared Error (MSE) by 39\% and improving the Structural Similarity (SSIM) from 0.926 to a near-perfect 0.988. This superior performance also generalizes to natural phenomena datasets. Our work not only delivers a new state-of-the-art (SOTA) model but, more importantly, provides the community with an invaluable data resource to drive future research in industrial AI.
\end{abstract}

\end{frontmatter}



\section{Introduction}
\label{sec:introduction}
\begin{figure}[htbp]
  \centering
  \includegraphics[width=0.45\textwidth,keepaspectratio]{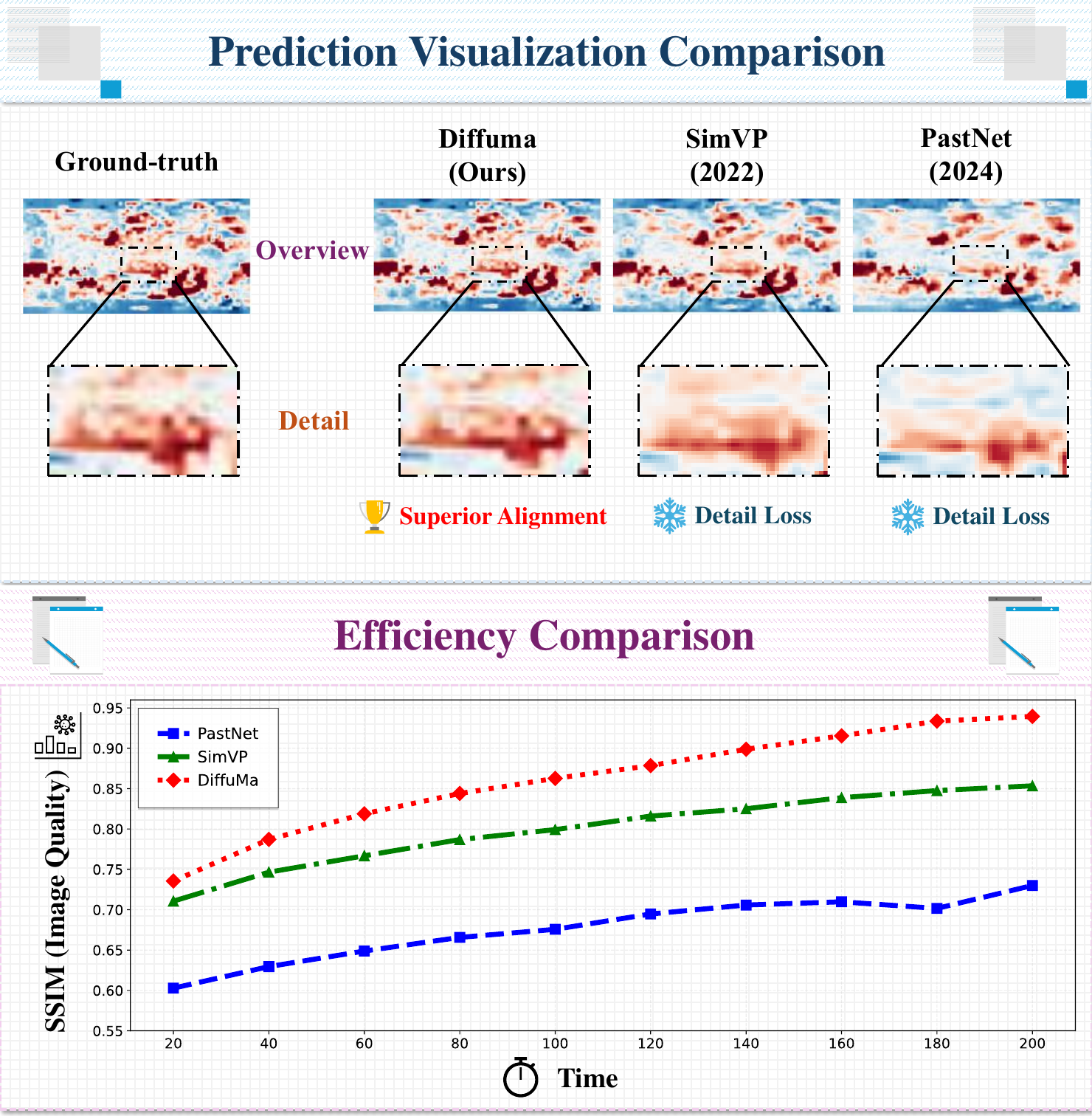}
    \caption{\textbf{Qualitative and Quantitative Comparison of Prediction Performance.}}
  \label{fig:introduction_main}
\end{figure}
Spatio-temporal video prediction, the task of generating future frames from a sequence of historical ones, constitutes a fundamental yet challenging problem in computer vision~\cite{wu2024pastnet, wu2024earthfarsser}. It plays an indispensable role in simulating macroscopic natural phenomena, such as weather forecasting and climate analysis \cite{li2025spatio,liu2023physics,li2024towards,lin2020self,wu2021graph,gao2023spatio,zhao2019t,aleissaee2023transformers,du2024multi}. In recent years, with the rise of Industry 4.0 and intelligent manufacturing, its value in microscopic industrial processes has become increasingly prominent. Particularly in high-precision domains like semiconductor manufacturing, accurately predicting the dynamic evolution of processes such as dicing and etching is paramount to achieving real-time process monitoring, early defect warning, and optimizing process parameters, directly impacting product yield and production efficiency. However, regardless of the scene's scale, generating long-term predictions that are both photorealistic and temporally coherent requires models to overcome two central challenges: first, effectively capturing complex, nonlinear temporal dynamics, and second, inhibiting the feature degradation caused by error accumulation, which manifests as blurring in predicted frames over time \cite{hochreiter1997long,cho2014learning,seo2018structured}.

To address the aforementioned challenges, the community has explored various approaches. Early architectures based on Recurrent Neural Networks (RNNs), such as ConvLSTM~\cite{shi2015convolutional}, pioneer the combination of convolution and sequence modeling \cite{wang2018eidetic,wu2019graph,jin2023spatio,kipf2016semi,yu2017spatio}. However, their inherent limitations in handling long-term dependencies, stemming from gradient issues, make it difficult for them to capture global evolutionary patterns \cite{sankar2020graph,cao2022network, bai2020adaptive, li2023dynamic}. Subsequently, Transformer-based models demonstrate advantages in modeling long-range relationships with their powerful self-attention mechanism, but their quadratic computational complexity with respect to sequence length makes their application to high-resolution, long-duration video data computationally prohibitive \cite{wu2021autoformer,gao2022earthformer,wu2022flowformer,lim2021temporal,shehzad2024graph}. More importantly, these discriminative models often fail to prevent the loss of detailed information during multi-step rollouts, leading to overly smooth predictions that lack real-world textures. Thus, on a \textbf{methodological level}, there is an urgent need for a new model architecture that can efficiently process long sequences like State Space Models (SSMs) while also possessing the powerful detail-restoration capabilities of generative models. Even more critically, on an \textbf{application level}, research in high-precision industrial scenarios has long been constrained by a severe lack of public, high-quality benchmark datasets. This \textbf{data gap} not only impedes the development and fair comparison of advanced algorithms but also creates a significant divide between academic innovation and real-world industrial needs.

To fill this twofold gap in methodology and application benchmarks, we make two core contributions in this work. \textbf{First, we construct and release the Chip Dicing Lane Dataset (CHDL)}, the \textbf{first} public temporal image dataset dedicated to the dynamic evolution of the semiconductor wafer dicing process. Captured via an industrial-grade, high-resolution vision system, its fine-grained textures and complex dynamics present an unprecedented challenge to existing models, while also providing an invaluable testbed for cutting-edge research in high-fidelity process modeling, micro-defect detection, and industrial digital twins. \textbf{Second, to effectively model the complex dynamics presented by CHDL, we propose DIFFUMA}, a novel \textbf{dual-path video prediction architecture}. The core idea of this architecture lies in decoupling the learning of spatio-temporal features: one path utilizes the advanced bidirectional Mamba (SSM) to efficiently capture \textbf{global, long-range temporal dependencies} with linear complexity, responsible for understanding the evolutionary logic of "what happens" \cite{gu2023mamba,gu2022combining,smith2022simplified}. The other parallel path acts as a \textbf{detail enhancer} by drawing inspiration from diffusion models; it reconstructs and sharpens spatial textures by learning a denoising task, ensuring the visual fidelity of "how it looks." Crucially, these two paths do not operate independently; the denoising process of the diffusion module is \textbf{precisely guided} by the temporal features extracted by the Mamba module, thereby achieving a deep synergy of spatio-temporal information and fundamentally addressing the blurring problem in long-term prediction \cite{karras2022elucidating,tashiro2021csdi,yang2024survey}. \textbf{\textit{The efficacy of our dual-path design is visually demonstrated in Figure~\ref{fig:introduction_main}. While leading methods like SimVP and PastNet suffer from significant detail loss and blurring over time, DIFFUMA's predictions maintain remarkable visual clarity and structural fidelity. This qualitative superiority is corroborated by quantitative metrics, where our model consistently achieves the highest SSIM (Image Quality) score across the entire prediction horizon, robustly demonstrating its exceptional capability for long-term, high-fidelity forecasting.}}

Our main contributions are summarized as follows:
\begin{itemize}
    \item \textbf{We propose CHDL, the first public video dataset for the semiconductor dicing process}, which not only provides a challenging new benchmark for high-precision industrial video prediction but also fills a critical data gap in this research area.
    \item \textbf{We design DIFFUMA, an innovative dual-path architecture} that, through the clever synergy of Mamba and a diffusion-enhanced mechanism, simultaneously addresses the two major challenges of efficient long-term dependency modeling and prediction feature degradation within a single model.
    \item \textbf{Extensive experiments on CHDL and other public datasets (e.g., WeatherBench) demonstrate} that DIFFUMA's performance significantly surpasses that of existing state-of-the-art baseline models. On CHDL in particular, it reduces the Mean Squared Error (MSE) by 39\% and improves the Structural Similarity (SSIM) from 0.926 to an impressive 0.988, fully validating the superior performance and strong generalization capability of our method.
\end{itemize}

The remainder of this paper is organized as follows: Section~\ref{sec:related_work} reviews related work, Section~\ref{sec:dataset} details the construction of the CHDL dataset, Section~\ref{sec:method} elaborates on the architecture of the DIFFUMA model, Section~\ref{sec:experiments} presents the experimental results and analysis, and finally, Section~\ref{sec:conclusion} provides a conclusion.

\section{Related Work}
\label{sec:related_work}

\subsection{Spatio-Temporal Prediction Models}

Spatio-temporal video prediction~\cite{wu2024pure, wu2024pastnet} aims to generate future visual content from historical frame sequences, and its model development has progressed through several stages. Early architectures based on Recurrent Neural Networks (RNNs), such as \textbf{ConvLSTM} \cite{shi2015convolutional}, pioneered the combination of convolution operations with sequential modeling units to effectively process spatio-temporal data. Subsequent models like \textbf{PredRNN} and \textbf{PredRNN++} \cite{wang2018predrnn++} further enhanced the modeling of complex dynamics by introducing spatio-temporal memory cells and gradient highway mechanisms. However, these methods are constrained by the serial computation paradigm of RNNs, often suffering from poor gradient propagation when handling long-term dependencies. This makes it difficult for the models to capture global evolutionary patterns across an entire sequence. Furthermore, errors accumulate frame-by-frame during multi-step prediction, leading to an inevitable loss of detail and blurring in long-term forecasts, a phenomenon known as \textbf{feature degradation}.

To overcome the challenge of long-term dependencies, architectures based on the \textbf{Transformer} were introduced to the video prediction domain~\cite{wu2025triton, wu2025turb}. Models such as variants of \textbf{ViT} \cite{dosovitskiy2020image} and \textbf{MAE-ST} \cite{he2022masked} leverage the self-attention mechanism to directly establish relationships between any two frames, showing significant advantages in capturing long-range spatio-temporal dependencies. Nevertheless, the core self-attention mechanism of the Transformer exhibits quadratic computational complexity with respect to the sequence length. This poses a substantial computational and memory burden when applied to high-resolution, long-sequence video data, severely limiting its practicality. This motivates us to explore a new paradigm that can efficiently capture global context while maintaining computational efficiency.

Recently, \textbf{State Space Models (SSMs)}, particularly their latest variant \textbf{Mamba} \cite{gu2023mamba}, have garnered significant attention as a powerful sequence modeling tool. Mamba achieves linear-time complexity for sequence modeling by introducing a selective scan mechanism, while demonstrating performance comparable or even superior to Transformers on long-sequence tasks like language modeling. Despite Mamba's tremendous success in other fields, its application in spatio-temporal video prediction remains in its infancy. Instead of a simple backbone replacement, we innovatively employ a \textbf{bidirectional Mamba structure} to more comprehensively encode temporal context and integrate it as the core temporal modeling engine within our unique \textbf{dual-path architecture}, specifically designed to address the distinct challenges of video prediction.

On another front, to fundamentally address the blurring issue in predictions, some works have begun to explore \textbf{generative models}. GAN-based methods can generate photorealistic images, but their training instability is well-known. More recently, \textbf{Diffusion Models} \cite{sohl2015deep} have gained prominence for their exceptional ability in high-fidelity image generation. However, a standard diffusion model requires a multi-step iterative denoising process to generate a single image, which is time-consuming and fails to meet the inference speed requirements of video prediction tasks. Diverging from these methods that require a full, multi-step generative process, we ingeniously \textbf{leverage the core denoising idea of diffusion models} and engineer it into an efficient, single-pass \textbf{"diffusion enhancement module."} It is not an independent generator but rather a feature enhancer guided by the Mamba pathway, specifically designed to restore and sharpen spatial details during prediction—a synergistic design that, to our knowledge, is unprecedented.

\subsection{Benchmarks for Video Prediction}

The advancement of the video prediction field is inseparable from the support of high-quality benchmark datasets. Currently, commonly used datasets primarily cover synthetic scenes, natural scenes, and scientific data. Synthetic datasets like \textbf{Moving MNIST} \cite{srivastava2015unsupervised} are often used for proof-of-concept validation due to their simplicity and controllability. In the realm of natural scenes, datasets such as \textbf{KTH Actions} \cite{schuldt2004recognizing} and \textbf{UCF101} \cite{soomro2012ucf101} focus on human action recognition, while \textbf{Cityscapes} \cite{cordts2016cityscapes} provides complex urban street-view videos. In the scientific domain, \textbf{WeatherBench} \cite{rasp2020weatherbench} (based on ERA5 reanalysis data) offers invaluable global meteorological data for weather forecasting research. Although these datasets have greatly propelled the progress of video prediction algorithms, they all fail to reflect the unique challenges present in high-precision industrial manufacturing processes, such as: \textbf{(1) repetitive and fine-grained microscopic textures; (2) extreme demands on prediction fidelity}, where any minor blurring or artifact could lead to the misjudgment of defects; and \textbf{(3) subtle yet critical dynamic changes}, like the formation of micro-cracks or the splashing of debris.

To bridge the gap between existing academic research and real-world industrial needs, we construct and propose the \textbf{CHDL (Chip Dicing Lane Dataset)}. To the best of our knowledge, this is the \textbf{first} public video benchmark dataset focused on the semiconductor wafer dicing process. CHDL is collected via an industrial-grade, high-resolution vision system, fully documenting the dynamic morphological evolution of the dicing lane during processing. It not only provides a new testbed with industrial realism and high-level challenges for video prediction models but, more importantly, offers invaluable data support for cutting-edge application research, including \textbf{intelligent industrial defect detection, online optimization of process parameters, and the development of high-fidelity digital twin systems}. Our goal is to empower academic innovation to play a greater role in solving practical industrial problems.

\section{The CHDL Dataset}
\label{sec:dataset}
\begin{figure*}[htbp]
  \centering
  \includegraphics[width=0.98\textwidth]{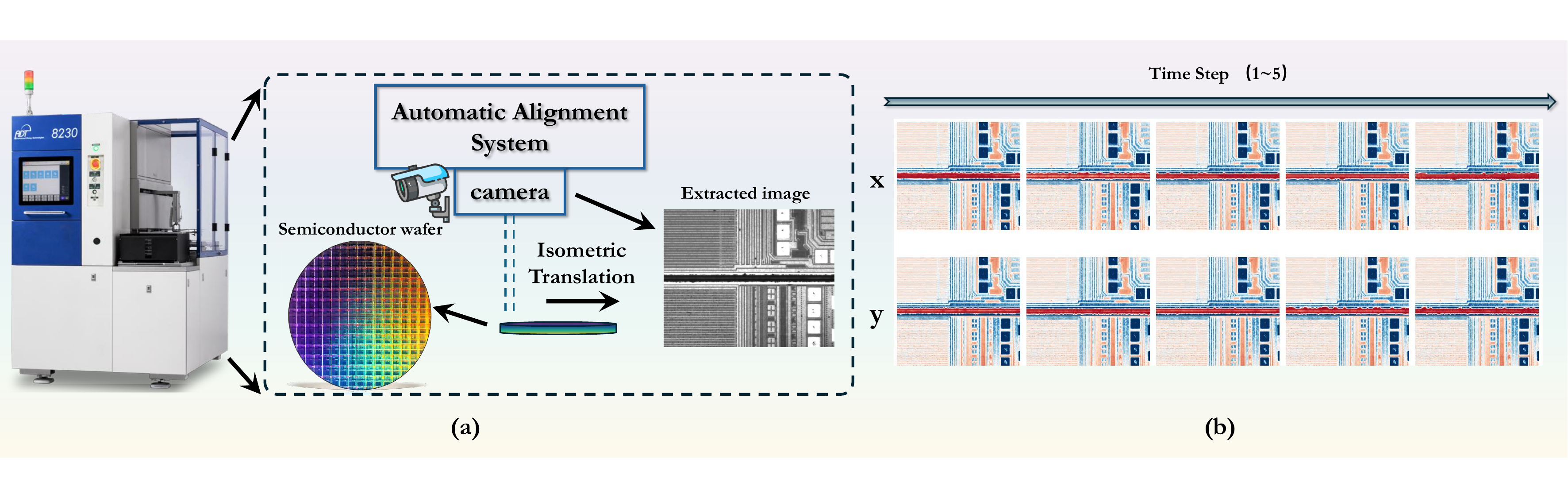} 
  \caption{\textbf{Overview of the CHDL Dataset Acquisition and Structure.} 
  (\textbf{a}) The data acquisition system, illustrating how the high-precision dicing machine (ADT-8230) is integrated with the camera in the automatic alignment system to capture microscopic images of the dicing lane during wafer translation.
  (\textbf{b}) An example of a CHDL data sample. Each sample consists of 10 consecutive frames, divided into an input sequence \textbf{x} (the first 5 frames) and a target prediction sequence \textbf{y} (the subsequent 5 frames), fully documenting the dynamic evolution of the dicing lane over time.}
  \label{fig:contribution_dataset}
\end{figure*}
To address the long-standing absence of benchmark data in the high-precision industrial video prediction domain, we construct and release the \textbf{CHDL (Chip Dicing Lane Dataset)}, the first public temporal image dataset dedicated to the dynamic evolution of the semiconductor wafer dicing process. CHDL not only establishes a highly challenging spatiotemporal benchmark for deep learning models but also provides high-quality data support for industrial defect detection and the development of digital twin systems. \textbf{Figure~\ref{fig:contribution_dataset}} provides a comprehensive overview of the CHDL dataset, from (a) the real-world industrial data acquisition system to (b) the structure of the resulting temporal image samples.

\subsection{Data Acquisition Process}

Our data acquisition system, as depicted in \textbf{Figure~\ref{fig:contribution_dataset}(a)}, is a synergistic integration of a high-precision dicing machine (ADT-8230) and a custom industrial vision subsystem. During the data acquisition process, a semiconductor wafer undergoes precise isometric translations on the stage. Concurrently, a high-resolution camera within the \textbf{Automatic Alignment System}, which is linked to the machine, captures microscopic images of the dicing lane in real-time.

To accurately record the progressive morphological evolution of the dicing lanes, we designed a rigorous acquisition protocol:
\begin{itemize}
    \item \textbf{Temporal Image Capture:} The automated alignment system moves along the dicing lanes at fixed intervals, acquiring 800$\times$600 grayscale images (C=1) at 10 equidistant positions with a fixed sampling rate of 5 fps. The entire process is conducted under strictly controlled illumination (2000$\pm$50 lux) to ensure data consistency.
    \item \textbf{Data Curation and Quality Control:} The raw data undergoes a meticulous curation process. Abnormal frames (approximately 1.3\% of the total), caused by mechanical vibrations or illumination fluctuations, are discarded and reconstructed via bilinear interpolation to ensure the integrity and smoothness of each sequence.
\end{itemize}

\subsection{Dataset Structure and Composition}

The structure of the CHDL dataset is designed to emulate real-world industrial prediction workflows. It is organized as a multidimensional tensor with a \textbf{BTCHW} (Batch, Time, Channel, Height, Width) structure, with the following key design features:
\begin{itemize}
    \item \textbf{Temporal Coherence:} As illustrated in \textbf{Figure~\ref{fig:contribution_dataset}(b)}, each data sample is a sequence pair containing 10 consecutive frames. The first 5 frames (labeled as \textbf{x}) serve as the model's input to capture the historical state, while the subsequent 5 frames (labeled as \textbf{y}) act as the prediction target. This paired input-output structure precisely defines the prediction task from past to future.
    \item \textbf{Spatial Integrity and Scale:} All images maintain their full 800$\times$600 pixel resolution, providing complete coverage of the critical regions of the dicing lane. After curation, the final dataset comprises \textbf{6,000} such input-target sample pairs, which are evenly split into 3,000 for training and 3,000 for validation, ensuring robust support for data-intensive models.
\end{itemize}

\subsection{Application Scenarios}

With its industrial realism and high fidelity, the CHDL dataset can robustly support the development and validation of cutting-edge algorithms for:
\begin{itemize}
    \item \textbf{Process Prediction:} Forecasting the future morphological evolution of dicing lanes based on preceding frame sequences.
    \item \textbf{Defect Detection:} Identifying micro-cracks, edge chipping, and other subtle defects by analyzing anomalies in spatio-temporal dynamics.
    \item \textbf{Digital Twins:} Generating high-fidelity training data for virtual dicing systems to accelerate the development and iteration of simulation models.
\end{itemize}

\section{Proposed Method}
\label{sec:method}
\begin{figure*}[!] 
  \centering
  \includegraphics[width=1\linewidth,keepaspectratio]{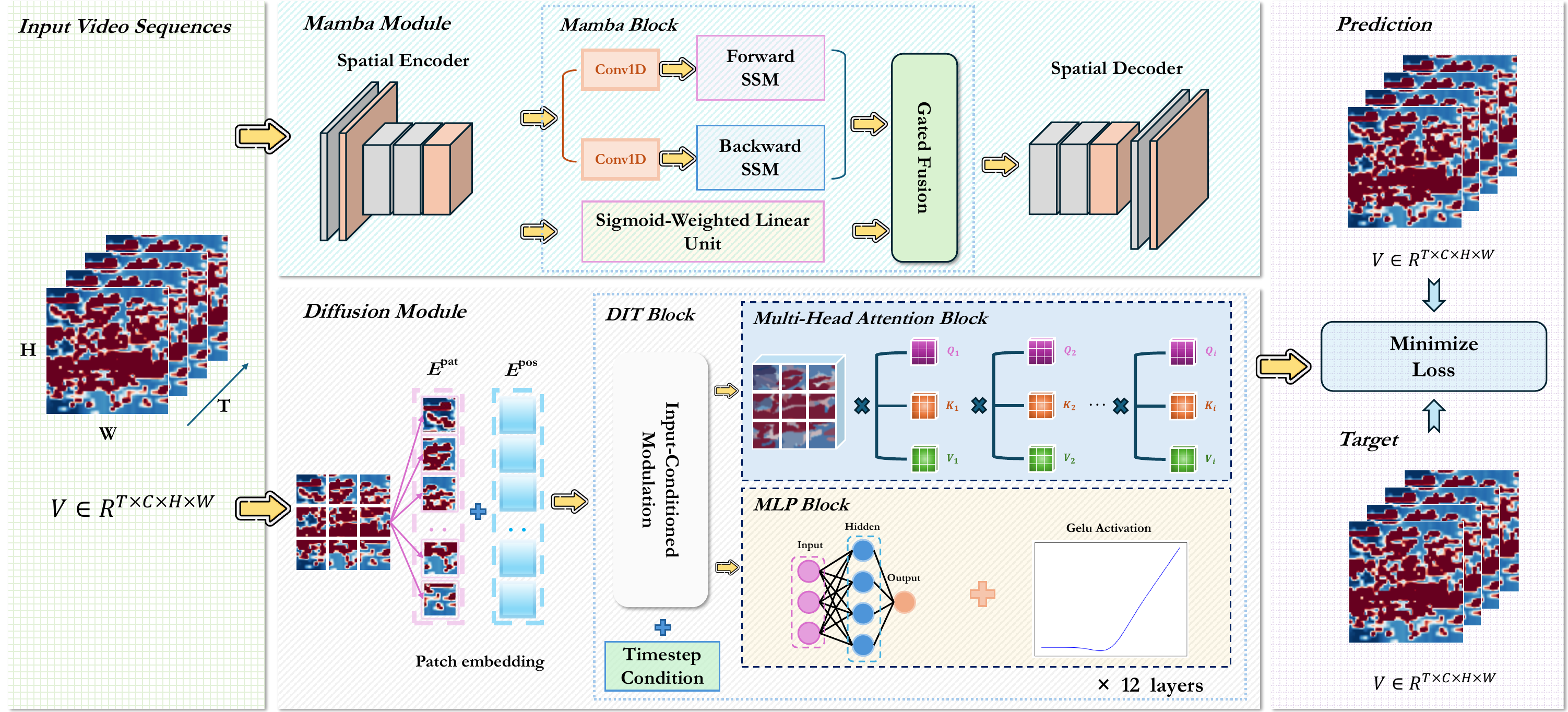}
  \caption{An overview of the DiffuMa. In this framework, the Mamba module models long-range temporal context by utilizing multi-layer bidirectional SSM and convolution, while the diffusion module enhances local detail retention through MLP block and multi-head attention block.}
  \label{fig:methodology_DiffuMa}
\end{figure*}

\subsection{Problem Formulation and Framework Overview}

The primary objective of spatio-temporal video prediction is to forecast a sequence of future frames based on a given sequence of past frames. Formally, let $\mathbf{X}_{1:T_{\text{in}}} = \{ \mathbf{x}_1, \mathbf{x}_2, \dots, \mathbf{x}_{T_{\text{in}}} \}$ be a sequence of $T_{\text{in}}$ input frames, where each frame $\mathbf{x}_t \in \mathbb{R}^{H \times W \times C}$ has a height $H$, width $W$, and $C$ channels. The goal is to generate a sequence of $T_{\text{out}}$ future frames, $\hat{\mathbf{Y}}_{T_{\text{in}}+1:T_{\text{in}}+T_{\text{out}}} = \{ \hat{\mathbf{y}}_{T_{\text{in}}+1}, \dots, \hat{\mathbf{y}}_{T_{\text{in}}+T_{\text{out}}} \}$, that is as close as possible to the ground-truth sequence $\mathbf{Y}_{T_{\text{in}}+1:T_{\text{in}}+T_{\text{out}}}$. This can be framed as learning a mapping function $\mathcal{F}: \mathbb{R}^{T_{\text{in}} \times H \times W \times C} \to \mathbb{R}^{T_{\text{out}} \times H \times W \times C}$.

To address the inherent challenges of modeling complex dynamics while preserving high-fidelity details, we propose \textbf{DIFFUMA}, an innovative dual-path architecture that decouples the learning of temporal evolution and spatial texture generation. As illustrated in Figure~\ref{fig:methodology_DiffuMa}, DIFFUMA comprises two synergistic core components: a \textbf{Mamba Module} and a \textbf{Diffusion Module}. Given an input video sequence, the Mamba Module serves as the temporal backbone, responsible for capturing long-range dependencies and global dynamic trends. It processes the entire input sequence $\mathbf{X}_{1:T_{\text{in}}}$ to produce a contextually rich latent representation, which forms a preliminary, dynamically coherent prediction. Concurrently, the Diffusion Module functions as a high-fidelity detail enhancer. It takes a noisy version of the input frames, conditioned on both a random timestep $t$ and, crucially, the global temporal context provided by the Mamba Module. Its objective is to learn a denoising function $\epsilon_\theta$ that predicts the noise added to the clean input, thereby restoring fine-grained spatial details. The final, high-fidelity prediction $\hat{\mathbf{Y}}$ is synthesized by fusing the outputs of both modules. The entire model is trained end-to-end by minimizing a loss function, such as the Mean Squared Error (MSE), between the prediction and the ground-truth target:
\begin{equation}
\tiny
\label{eq:total_loss}
\mathcal{L}_{\text{total}} = \mathbb{E}_{\mathbf{X}, \mathbf{Y}, \epsilon \sim \mathcal{N}(0, \mathbf{I}), t} \left[ \left\| \epsilon - \mathcal{F}(\mathbf{X}_{1:T_{\text{in}}}, \sqrt{\bar{\alpha}_t}\mathbf{X}_{1:T_{\text{in}}} + \sqrt{1-\bar{\alpha}_t}\epsilon, t) \right\|^2_2 \right]
\end{equation}
where $\mathcal{F}$ represents the entire DIFFUMA network, $\epsilon$ is the sampled noise, and $\bar{\alpha}_t$ is the noise schedule from diffusion theory. This formulation elegantly integrates a generative, detail-oriented objective within a predictive, dynamics-focused framework. The subsequent sections will elaborate on the specific architectural designs of the Mamba and Diffusion modules.

\subsection{Mamba Module: Efficient Temporal Dynamics Modeling}

The Mamba Module serves as the core of DIFFUMA for modeling dynamics, designed to efficiently capture complex, long-range temporal dependencies throughout the video sequence. As depicted in Figure~\ref{fig:methodology_DiffuMa}, this module consists of a spatial encoder, a series of Bidirectional Mamba Blocks, and a spatial decoder.

\paragraph{Spatial Encoder.}
Given an input video sequence $\mathbf{X} \in \mathbb{R}^{T \times C \times H \times W}$, we first transform it into a temporal feature sequence using a spatial encoder. This encoder, composed of several 2D convolutional layers, processes each frame independently to compress high-dimensional spatial information into a compact latent representation. This process can be formulated as:
\begin{equation}
\mathbf{Z}^{(0)} = \text{SpatialEncoder}(\mathbf{X})
\end{equation}
where $\mathbf{Z}^{(0)} \in \mathbb{R}^{T \times D}$ is the initial feature sequence fed into the Mamba Blocks, and $D$ is the feature dimension. This step effectively reduces the computational complexity of subsequent operations and extracts features most relevant to the dynamic changes.

\paragraph{Bidirectional Mamba Block.}
The encoded feature sequence $\mathbf{Z}^{(0)}$ is then processed by a stack of $L$ Bidirectional Mamba Blocks for deep temporal information interaction. Each Mamba Block is the heart of this module, synergistically processing temporal information through a carefully designed multi-branch structure. For the $l$-th Mamba block, its input is $\mathbf{Z}^{(l-1)}$, and its internal workflow is as follows:

First, the input features $\mathbf{Z}^{(l-1)}$ are fed into three parallel branches. Two of these branches are dedicated to capturing temporal dynamics. They each pass through a 1D convolution (Conv1D) to enhance the perception of local temporal context before being channeled into a \textbf{Forward State Space Model (SSM)} and a \textbf{Backward State Space Model (SSM)}, respectively. The core idea of an SSM is to summarize historical information via a latent state $\mathbf{h}_t \in \mathbb{R}^{N}$, with its state update being abstractly represented as:
\begin{equation}
\mathbf{h}_t = \mathbf{\bar{A}}\mathbf{h}_{t-1} + \mathbf{\bar{B}}\mathbf{z}_t, \quad \mathbf{o}_t = \mathbf{C}\mathbf{h}_t
\end{equation}
where $\mathbf{\bar{A}}, \mathbf{\bar{B}}, \mathbf{C}$ are learnable matrices obtained by discretizing continuous SSM parameters. The Forward SSM processes the sequence in the order $t=1, \dots, T$, while the Backward SSM processes it in reverse order, $t=T, \dots, 1$. This bidirectional design enables the model to integrate both past and future contexts at each timestep, thereby forming a holistic understanding of the temporal dynamics. We fuse the outputs of the forward and backward paths (e.g., via addition or concatenation) to obtain the bidirectional dynamic features $\mathbf{O}_{\text{bi-ssm}}$.

Parallel to the SSM branches is a \textbf{gating branch}, which implements Mamba's selection mechanism. The input features $\mathbf{Z}^{(l-1)}$ undergo a linear transformation and are then passed through a Sigmoid-Weighted Linear Unit to generate a gating signal $\mathbf{G} \in \mathbb{R}^{T \times D}$.

Finally, through \textbf{Gated Fusion}, the dynamic features extracted by the SSM branches are element-wise multiplied with the selective signal generated by the gating branch, yielding the final output of the Mamba block:
\begin{equation}
\mathbf{Z}^{(l)} = \text{GatedFusion}(\mathbf{O}_{\text{bi-ssm}}, \mathbf{G}) = \mathbf{O}_{\text{bi-ssm}} \odot \mathbf{G}
\end{equation}
This gating mechanism allows the model to dynamically and selectively amplify or suppress information flow, which is crucial for modeling complex, nonlinear dynamics.

\paragraph{Spatial Decoder.}
After being processed by $L$ layers of Mamba blocks, the final feature sequence $\mathbf{Z}^{(L)}$ contains rich, long-range, and context-aware spatio-temporal information. We feed this sequence into a spatial decoder, composed of several transposed convolutional layers, to upsample the features back to the original image space. This generates a preliminary, dynamically coherent prediction $\hat{\mathbf{Y}}_{\text{mamba}} \in \mathbb{R}^{T_{\text{out}} \times C \times H \times W}$. This result captures the macroscopic motion and structural changes of the video and serves as crucial guidance for the subsequent Diffusion Module.

\subsection{Diffusion Module: High-Fidelity Spatial Detail Enhancement}

A prevalent issue in traditional video prediction models is feature degradation during long-term forecasting, which leads to blurry outputs. To fundamentally address this problem, we design a \textbf{Diffusion Module}. Unlike standard generative diffusion models that require multi-step iterative sampling, our module is an efficient, \textbf{single-pass feed-forward denoiser}. Its core task is to restore and enhance fine-grained spatial details under the guidance of the global dynamics provided by the Mamba Module. The architecture of this module is heavily inspired by the \textbf{Diffusion Transformer (DiT)}.

\paragraph{Noising Process and Conditioning.}
For a given input video sequence $\mathbf{X}$, we first sample a random timestep $t$ from a uniform distribution $\mathcal{U}(1, T_{\text{diff}})$ and add Gaussian noise according to a predefined noise schedule $\bar{\alpha}_t$, yielding a noised input $\mathbf{x}_t' = \sqrt{\bar{\alpha}_t}\mathbf{x} + \sqrt{1-\bar{\alpha}_t}\epsilon$, where $\epsilon \sim \mathcal{N}(0, \mathbf{I})$. The objective of our diffusion module is to train a network $\epsilon_\theta$ to predict the added noise $\epsilon$.

To make the denoising process controllable and precise, we inject critical conditional information into the model:
\begin{itemize}
    \item \textbf{Timestep Condition:} The timestep $t$ is converted into an embedding vector $\mathbf{c}_{\text{time}} = \text{Embed}(t)$, which explicitly informs the network of the current noise level.
    \item \textbf{Input-Conditioned Modulation:} This is the core of our synergistic dual-path design. We leverage the feature representation $\hat{\mathbf{Y}}_{\text{mamba}}$, which is rich in global dynamic information from the \textbf{Mamba Module}, as a second condition $\mathbf{c}_{\text{context}}$. These two conditional vectors are fused and used to modulate the normalization layers within the DiT blocks, thereby achieving precise guidance over the denoising process. This design ensures that the restoration of spatial details is \textbf{consistent with the overall motion patterns}.
\end{itemize}

\paragraph{DiT Block.}
The DiT Block is the main body of the Diffusion Module. First, the noised image $\mathbf{x}_t'$ is partitioned into a series of non-overlapping patches, which are then linearly projected into Patch Embeddings $\mathbf{E}^{\text{pat}}$. After adding a learnable positional encoding $\mathbf{E}^{\text{pos}}$, the resulting feature sequence is fed into a Transformer structure composed of 12 DiT blocks.

As shown in Figure~\ref{fig:methodology_DiffuMa}, each DiT block consists of a \textbf{Multi-Head Self-Attention (MHSA) block} and an \textbf{MLP block}, combined via residual connections. The MHSA block is responsible for capturing long-range spatial dependencies among different image patches, thus integrating global information to infer local details. The MLP block performs nonlinear feature transformations. Within each DiT block, the injected conditional information $(\mathbf{c}_{\text{time}}, \mathbf{c}_{\text{context}})$ dynamically modulates the features through mechanisms like Adaptive Layer Normalization (AdaLN), applying an affine transformation to the outputs of the MHSA and MLP blocks. The overall process can be summarized as:
\begin{align}
\mathbf{z}_{l}' &= \text{MHSA}(\text{AdaLN}(\mathbf{z}_{l-1}'; \mathbf{c})) + \mathbf{z}_{l-1}' \\
\mathbf{z}_{l} &= \text{MLP}(\text{AdaLN}(\mathbf{z}_{l}'; \mathbf{c})) + \mathbf{z}_{l}'
\end{align}
where $\mathbf{z}_{l-1}'$ and $\mathbf{z}_{l}$ are the input to and output from the $l$-th block, respectively, and $\mathbf{c}$ is the fused conditional vector. By stacking multiple DiT blocks, the model learns a powerful capability to restore fine-grained textures from noisy images.

\paragraph{Output and Fusion.}
The final output of the Diffusion Module is the predicted noise $\hat{\epsilon} = \epsilon_\theta(\mathbf{x}_t', \mathbf{c}_{\text{time}}, \mathbf{c}_{\text{context}})$. During inference, we set the timestep $t=0$ (i.e., no noise), where the output of the Diffusion Module can be interpreted as a \textbf{detail-enhancement residual} $\Delta \mathbf{X}$ for the input image. We fuse this residual with the preliminary prediction from the Mamba Module to obtain the final, high-fidelity predicted video:
\begin{equation}
\hat{\mathbf{Y}} = \hat{\mathbf{Y}}_{\text{mamba}} + \Delta \mathbf{X}
\end{equation}
This fusion strategy allows the predictions of DIFFUMA to retain the temporal coherence provided by the Mamba Module while being enriched with the fine spatial details restored by the Diffusion Module, thus generating future frames that are both sharp and dynamically accurate.

\subsection{Training Objective and Implementation Details}

\paragraph{Training Objective.}
Our DIFFUMA model is designed as an end-to-end system, where the parameters of the entire network, including both the Mamba and Diffusion modules, are jointly optimized through a unified loss function. Our primary training objective is the denoising task of the Diffusion Module, which minimizes the Mean Squared Error (MSE) between the predicted noise and the ground-truth added noise:
\begin{equation}
\mathcal{L}_{\text{diff}} = \mathbb{E}_{\mathbf{x}, t, \epsilon} \left[ \left\| \epsilon - \epsilon_\theta(\sqrt{\bar{\alpha}_t}\mathbf{x} + \sqrt{1-\bar{\alpha}_t}\epsilon, t, \mathbf{c}_{\text{context}}) \right\|^2_2 \right]
\end{equation}
The design of this loss function is particularly strategic. For the denoising network $\epsilon_\theta$ to accurately predict the noise $\epsilon$, it must effectively leverage the guidance condition $\mathbf{c}_{\text{context}}$. Since $\mathbf{c}_{\text{context}}$ is directly derived from the Mamba Module, this objective implicitly compels the Mamba Module to learn a precise and meaningful representation of the video's dynamics. In other words, optimizing the denoising task concurrently optimizes the modeling of temporal dynamics.

To further ensure that the final prediction aligns with the ground-truth at the pixel level, we also incorporate a direct reconstruction loss. We use the L1 loss to measure the discrepancy between the final prediction $\hat{\mathbf{Y}}$ and the ground-truth target $\mathbf{Y}$:
\begin{equation}
\mathcal{L}_{\text{recon}} = \mathbb{E}_{\mathbf{X}, \mathbf{Y}} \left[ \|\hat{\mathbf{Y}} - \mathbf{Y}\|_1 \right]
\end{equation}
The final total loss is a weighted sum of these two loss terms:
\begin{equation}
\mathcal{L}_{\text{total}} = \mathcal{L}_{\text{diff}} + \lambda \mathcal{L}_{\text{recon}}
\end{equation}
where $\lambda$ is a hyperparameter that balances the implicit supervision for dynamic modeling with the explicit supervision for the final output.

\section{Experiments}
\label{sec:experiments}
\subsection{Experimental Setup}
\subsubsection{Datasets}
To comprehensively demonstrate the versatility of the model in addressing diverse complex problems, we conducted experimental evaluations across five distinct datasets. The datasets employed in this study are categorized into two groups based on their research objectives: the Weatherbench dataset designed for investigating natural phenomena, and the CHDL dataset specifically developed by this study to examine variations in dicing groove. Table~\ref{tab:datasets_details} provides a detailed summary of the datasets used in this research.

\subsubsection{Evaluation Metrics}
To assess the quality of predictions, we employ three evaluation metrics: Mean Squared Error (MSE), Mean Absolute Error (MAE), and Structural Similarity Index Measure(SSIM). Lower values of MSE and MAE, along with higher values of SSIM, indicate better model performance.

\subsubsection{Baseline Models}
We conducted a comprehensive evaluation by comparing DiffuMa with multiple baseline models on both non-natural phenomena and natural phenomena datasets. This includes competitive RNN architectures such as ConvLSTM, PredRNN-v1, PredRNN-v2, and E3D-LSTM. Additionally, we evaluated SimVP, a state-of-the-art CNN architecture designed for non-natural phenomena datasets, as well as PastNet, a versatile model that performs well on both natural and non-natural phenomena datasets.

\subsubsection{Experimental Environment}
All experiments are conducted on a machine equipped with 7 Nvidia GeForce RTX 3090 GPUs. The models are implemented using PyTorch. For fairness, all baseline models are trained and evaluated under the same conditions.

\begin{table*}[t]
\centering
\caption{Presents the statistical details of the datasets utilized in our experimental evaluations. For each dataset, $N_{train}$ and $N_{test}$ denote the number of samples in the training and test sets, respectively. The dimensions of each image frame are represented as $(C, H, W)$, while $T$ and $K$ indicate the lengths of the input and prediction sequences, respectively.}
\begin{tabular}{llcccccc}
\toprule
\textbf{Dataset} & \textbf{Type} & \textbf{$N_{train}$} & \textbf{$N_{test}$} & \textbf{$(C, H, W)$} & \textbf{$T$} & \textbf{$K$} \\
\midrule
Weatherbench & Cloud\_cover & 2300 & 657 & (1, 32, 64) & 12 & 12 \\
 & Component\_of\_wind & 2300 & 657 & (1, 32, 64) & 12 & 12 \\
 & Humidity & 2300 & 657 & (1, 32, 64) & 12 & 12 \\
 & Temperature & 2300 & 657 & (1, 32, 64) & 12 & 12 \\
\midrule
CHDL &  & 4874 & 1216 & (1, 600, 800) & 5 & 1 \\
\bottomrule
\end{tabular}
\label{tab:datasets_details}
\end{table*}

\begin{table*}[!t]
\centering
\caption{Comparison of the quantitative prediction results between DiffuMa and baseline models on both non-natural phenomenon datasets and natural phenomenon datasets. The evaluation metrics selected for this study include Mean Squared Error (MSE), Mean Absolute Error (MAE), and Structural Similarity (SSIM). For MSE and MAE, lower values indicate better performance, while for SSIM, higher values indicate better performance.}
\label{tab:performance_comparison}
\resizebox{\textwidth}{!}{ 
\begin{tabular}{lccccccccccccccc}
\toprule
 & \multicolumn{12}{c}{\textbf{Weatherbench}} & \multicolumn{3}{c}{\textbf{CHDL}} \\
\cmidrule(lr){2-13} \cmidrule(lr){14-16}
\textbf{Model} & \multicolumn{3}{c}{\textbf{Cloud\_cover}} & \multicolumn{3}{c}{\textbf{Component\_of\_wind}} & \multicolumn{3}{c}{\textbf{Humidity}} & \multicolumn{3}{c}{\textbf{Temperature}} & \textbf{} & \textbf{} & \textbf{} \\
 & \textbf{MSE} & \textbf{MAE} & \textbf{SSIM} & \textbf{MSE} & \textbf{MAE} & \textbf{SSIM} & \textbf{MSE} & \textbf{MAE} & \textbf{SSIM} & \textbf{MSE} & \textbf{MAE} & \textbf{SSIM} & \textbf{MSE} & \textbf{MAE} & \textbf{SSIM} \\
\midrule
Convlstm & 0.4651 & 0.5133 & 0.4175 & 0.1564 & 0.2904 & 0.7976 & 0.2536 & 0.3689 & 0.6476 & 0.2519 & 0.3664 & 0.6597 & 0.3172 & 0.5641 & 0.6631 \\
Predrnn-v1 & 0.3730 & 0.3962 & 0.4849 & 0.1255 & 0.2238 & 0.8270 & 0.2039 & 0.2849 & 0.7524 & 0.2021 & 0.2828 & 0.7654 & 0.2554 & 0.5046 & 0.7697 \\
E3D LSTM & 0.3183 & 0.3429 & 0.5089 & 0.1068 & 0.1943 & 0.8706 & 0.1742 & 0.2468 & 0.7877 & 0.1729 & 0.2450 & 0.8031 & 0.2197 & 0.4694 & 0.8077 \\
PastNet & 0.2160 & 0.2840 & 0.7150 & 0.1216 & 0.2624 & 0.8398 & 0.1595 & 0.3062 & 0.7160 & 0.0080 & 0.0647 & 0.9498 & 0.1528 & 0.3916 & 0.7318 \\
Predrnn-v2 & 0.1891 & 0.2029 & 0.5220 & 0.0643 & 0.1551 & 0.8977 & 0.1034 & 0.1461 & 0.8104 & 0.1023 & 0.1443 & 0.8256 & 0.1315 & 0.3624 & 0.8292 \\
SimVP & 0.1044 & 0.2248 & 0.8209 & 0.0380 & 0.1489 & 0.9436 & 0.0495 & 0.1717 & 0.9317 & 0.0030 & 0.0405 & 0.9854 & 0.0528 & 0.2297 & 0.8350 \\
DiffuMa & 0.0513 & 0.1505 & 0.8899 & 0.0221 & 0.1154 & 0.9542 & 0.0214 & 0.1131 & 0.9454 & 0.0028 & 0.0413 & 0.9878 & 0.0371 & 0.1925 & 0.9254 \\
\bottomrule
\vspace{13pt} 
\end{tabular}
}
\end{table*}

\begin{figure*}[!h] 
  \centering
  \includegraphics[width=0.98\linewidth,keepaspectratio]{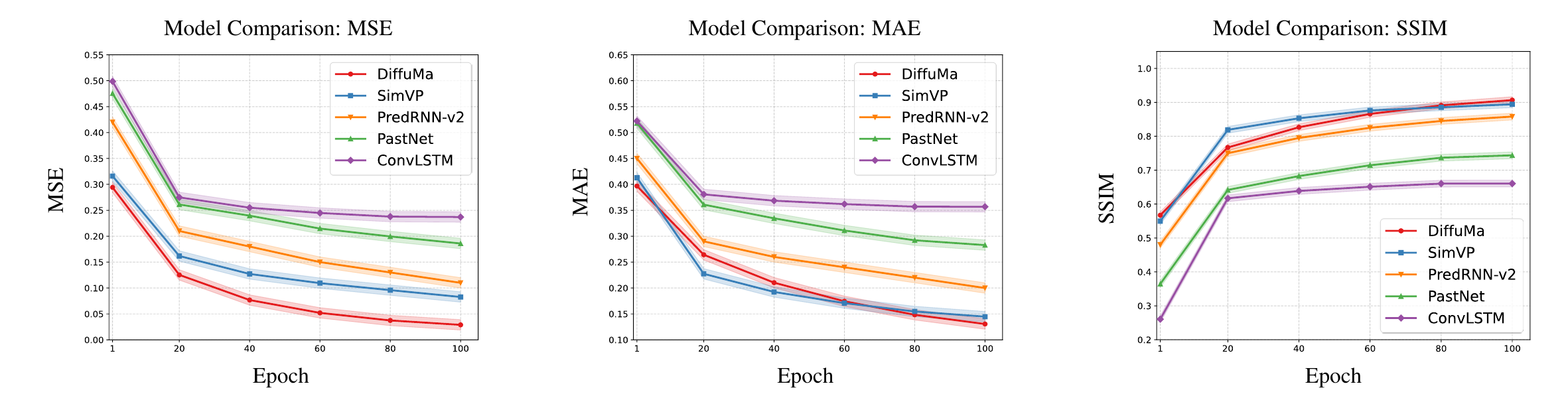}
  \caption{Performance Comparison: A Comprehensive Analysis of MSE, MAE and SSIM Trends Across Epochs}
  \label{fig:experiment_efficiency}
  \vspace{16pt} 
\end{figure*}

\subsection{Main Results}
Table~\ref{tab:performance_comparison} demonstrates that DiffuMa outperforms other models across both natural and non-natural phenomenon datasets. Specifically, on the natural phenomenon dataset WeatherBench, DiffuMa achieves state-of-the-art performance with MSE and MAE metrics that are 39\% and 21\% lower on average than those of SimVP (the most competitive baseline), respectively. Furthermore, it attains superior SSIM scores compared to other approaches, indicating enhanced capability in capturing spatiotemporal dynamics for video prediction. When evaluated on the non-natural phenomenon dataset CHDL, DiffuMa maintains its leading position by reducing the MSE and MAE metrics by 28\% and 37\%, respectively, compared to SimVP, while continuing to exhibit the highest SSIM values. These comprehensive experimental results substantiate that DiffuMa establishes statistically significant advantages over all baseline methods across every evaluation metric.

\begin{figure*}[htbp] 
  \centering
  \includegraphics[width=0.98\linewidth,keepaspectratio]{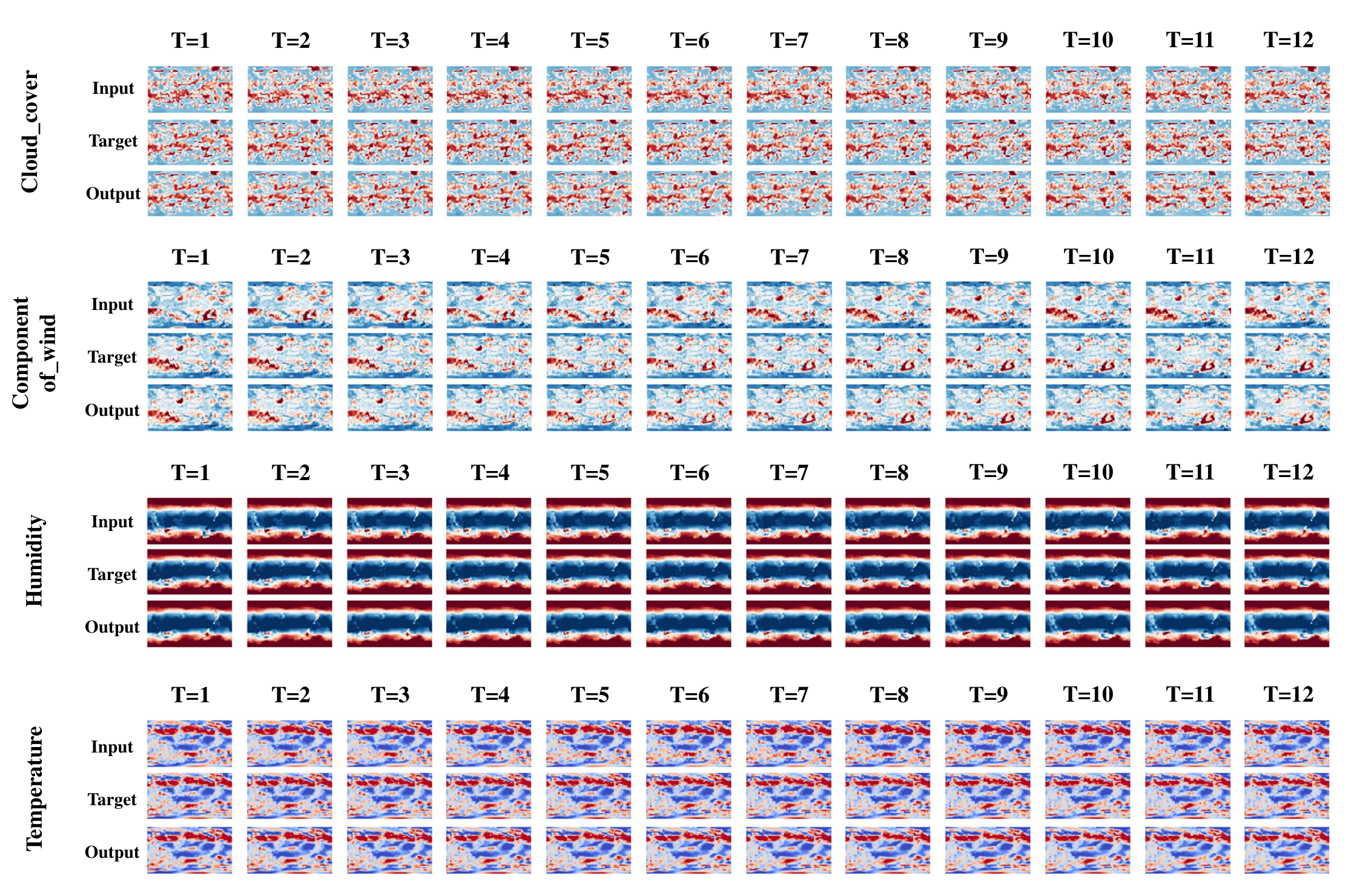}
  \vspace{-5pt}  
  \caption{Predictive Performance of DiffuMa on WeatherBench}
  \label{fig:experiment_weatherbench_main} 
\end{figure*}

\begin{figure*}[htbp] 
  \centering
  \includegraphics[width=0.98\linewidth,keepaspectratio]{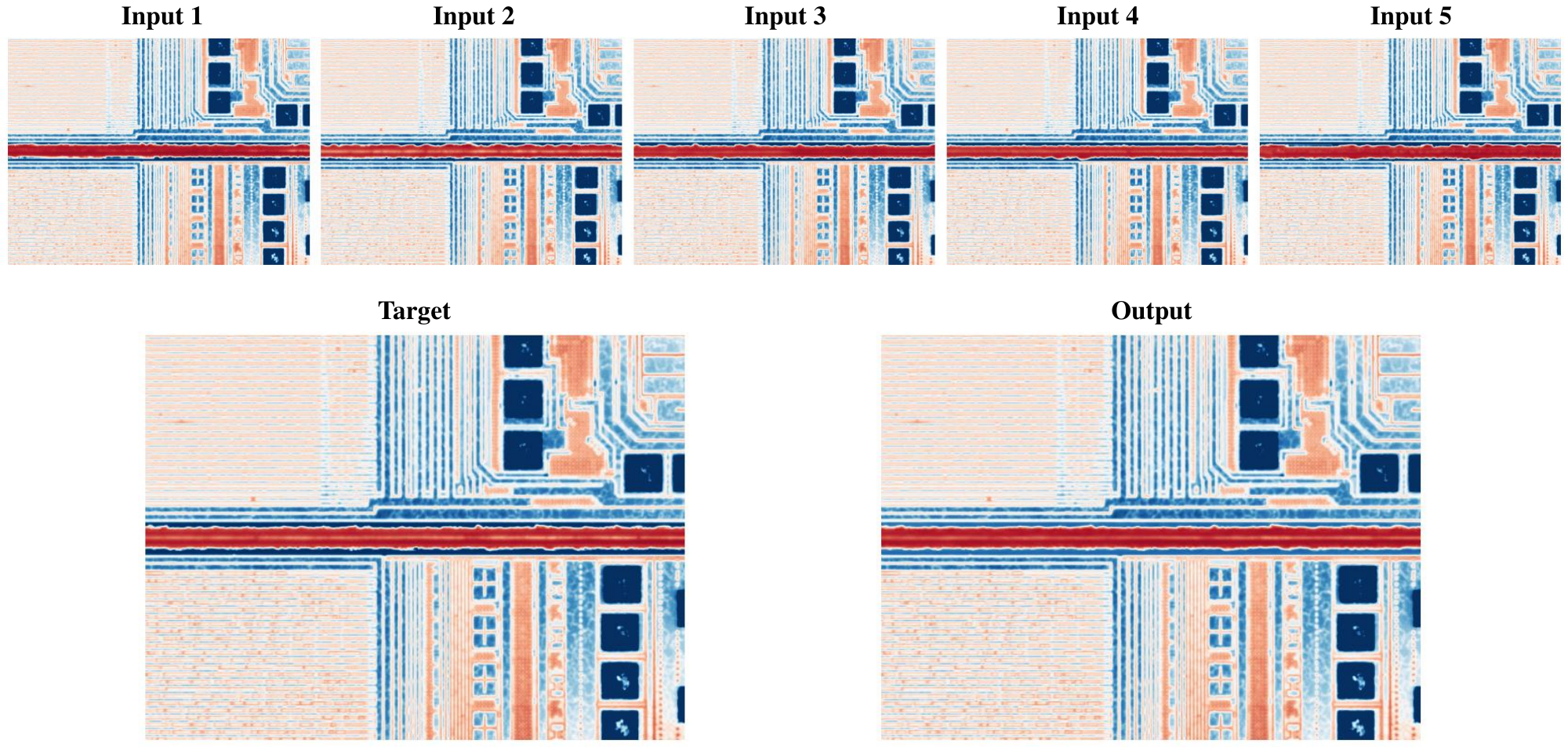}
  \caption{Predictive Performance of DiffuMa on CHDL}
  \label{fig:experiment_CHDL_main}
\end{figure*}
\subsection{Visualization of Prediction Results}
As illustrated in Figure~\ref{fig:experiment_weatherbench_main} and Figure~\ref{fig:experiment_CHDL_main}, we present the prediction results of DiffuMa across various datasets to demonstrate its predictive capabilities. In the prediction visualization for WeatherBench, the first row displays the input data, the second row shows the ground-truth, and the third row presents the model's predictions. For CHDL, the top row displays input sequences while the bottom row presents comparative results: ground-truth images (left) and model-predicted outputs (right). All visualizations maintain identical dimensions (800×600 pixels) in their original resolution. Notably, the ground-truth and predicted images have been proportionally upscaled using bicubic interpolation for enhanced visual clarity, without altering their inherent spatial relationships or introducing artifacts.

DiffuMa demonstrates exceptional predictive performance on benchmark datasets. Quantitative evaluations reveal that the structural similarity index (SSIM) between the generated images and the ground-truth targets reaches 0.926±0.018 on CHDL, and achieves a peak value of 0.988 on Weatherbench. Visual comparative analyses further confirm that the synthesized results exhibit high consistency with authentic targets in texture details, edge sharpness, and global structural features. Notably, DiffuMa significantly outperforms conventional baseline methods in noise suppression and detail preservation capabilities under complex scenarios. These results indicate that the proposed model effectively mitigates information loss during image degradation processes.

\begin{table*}[htbp]
\centering
\caption{MSE and MAE Values Across Epochs for Different Models (MSE $\downarrow$, MAE $\downarrow$)}
\label{tab:Epochs_mae_mse}
\begin{tabular}{l|rrrrrrrrrrrrrr}
\hline
\textbf{Epoch} & \multicolumn{2}{c}{10} & \multicolumn{2}{c}{20} & \multicolumn{2}{c}{40} & \multicolumn{2}{c}{60} & \multicolumn{2}{c}{80} & \multicolumn{2}{c}{100} \\
\hline  
\textbf{Model} & \textbf{MSE} & \textbf{MAE} & \textbf{MSE} & \textbf{MAE} & \textbf{MSE} & \textbf{MAE} & \textbf{MSE} & \textbf{MAE} & \textbf{MSE} & \textbf{MAE} & \textbf{MSE} & \textbf{MAE} \\
\hline
Convlstm      & 0.2860 & 0.3876 & 0.2632 & 0.3736 & 0.2515 & 0.3663 & 0.2470 & 0.3635 & 0.2404 & 0.3590 & 0.2382 & 0.3576 \\
Predrnn-v1    & 0.2753 & 0.3824 & 0.2547 & 0.3703 & 0.2435 & 0.3617 & 0.2352 & 0.3558 & 0.2281 & 0.3505 & 0.2223 & 0.3462 \\
E3D LSTM      & 0.2728 & 0.3789 & 0.2516 & 0.3657 & 0.2379 & 0.3574 & 0.2263 & 0.3485 & 0.2174 & 0.3417 & 0.2095 & 0.3358 \\
PastNet       & 0.2699 & 0.3812 & 0.2495 & 0.3673 & 0.2277 & 0.3536 & 0.2098 & 0.3417 & 0.1970 & 0.3329 & 0.1821 & 0.3214 \\
Predrnn-v2    & 0.1983 & 0.3355 & 0.1836 & 0.3209 & 0.1708 & 0.3057 & 0.1594 & 0.2926 & 0.1485 & 0.2808 & 0.1387 & 0.2689 \\
SimVP         & 0.1778 & 0.3133 & 0.1447 & 0.2862 & 0.1249 & 0.2677 & 0.1119 & 0.2546 & 0.0979 & 0.2391 & 0.0876 & 0.2269 \\
DiffuMa       & 0.1579 & 0.2901 & 0.1030 & 0.2419 & 0.0753 & 0.2085 & 0.0573 & 0.1828 & 0.0457 & 0.1639 & 0.0380 & 0.1494 \\
\hline
\end{tabular}
\end{table*}

\begin{table*}[htbp]
\centering
\caption{SSIM Values Across Epochs for Different Models (SSIM $\uparrow$)}
\label{tab:ssim_full}
\begin{tabular}{l|cccccccccccc}
\hline
\textbf{Epoch} & \textbf{1} & \textbf{10} & \textbf{20} & \textbf{30} & \textbf{40} & \textbf{50} & \textbf{60} & \textbf{70} & \textbf{80} & \textbf{90} & \textbf{100} & \textbf{110} \\
\hline
\textbf{Model} & \multicolumn{12}{c}{SSIM} \\
\hline
Convlstm      & 0.4048 & 0.5847 & 0.6160 & 0.6302 & 0.6369 & 0.6462 & 0.6474 & 0.6550 & 0.6557 & 0.6594 & 0.6620 & 0.6642\\
Predrnn-v1    & 0.4423 & 0.5908 & 0.6185 & 0.6337 & 0.6452 & 0.6573 & 0.6641 & 0.6708 & 0.6729 & 0.6850 & 0.6923 & 0.7007\\
E3D LSTM      & 0.4239 & 0.5936 & 0.6192 & 0.6354 & 0.6498 & 0.6629 & 0.6725 & 0.6774 & 0.6810 & 0.6997 & 0.7068 & 0.7081\\
PastNet       & 0.3609 & 0.5969 & 0.6201 & 0.6368 & 0.6541 & 0.6686 & 0.6800 & 0.6843 & 0.6893 & 0.7110 & 0.7031 & 0.7325\\
Predrnn-v2    & 0.4813 & 0.6532 & 0.6784 & 0.7059 & 0.7217 & 0.7395 & 0.7512 & 0.7640 & 0.7781 & 0.7924 & 0.8036 & 0.8103\\
SimVP         & 0.5519 & 0.7106 & 0.7402 & 0.7655 & 0.7835 & 0.7909 & 0.8027 & 0.8248 & 0.8268 & 0.8388 & 0.8478 & 0.8492\\
DiffuMa       & 0.5839 & 0.7228 & 0.7746 & 0.8106 & 0.8325 & 0.8508 & 0.8732 & 0.8846 & 0.8947 & 0.9022 & 0.9121 & 0.9207\\
\hline
\end{tabular}
\end{table*}

\begin{table*}[!b]
\centering
\caption{Performance comparison of different methods across prediction lengths (MSE $\downarrow$, MAE $\downarrow$, SSIM $\uparrow$)}
\label{tab:PL_mae_mse_ssim}
\begin{tabular}{l*{4}{ccc}}
\toprule
\multirow{2}{*}{Model} & 
\multicolumn{3}{c}{PL 3} & 
\multicolumn{3}{c}{PL 6} & 
\multicolumn{3}{c}{PL 9} & 
\multicolumn{3}{c}{PL 12} \\
\cmidrule(lr){2-4} \cmidrule(lr){5-7} \cmidrule(lr){8-10} \cmidrule(lr){11-13}
& \textbf{MSE} & \textbf{MAE} & \textbf{SSIM} & \textbf{MSE} & \textbf{MAE} & \textbf{SSIM} & \textbf{MSE} & \textbf{MAE} & \textbf{SSIM} & \textbf{MSE} & \textbf{MAE} & \textbf{SSIM} \\
\midrule
DiffuMa & 0.0095 & 0.0748 & 0.9434 & 0.0159 & 0.0964 & 0.9277 & 0.0225 & 0.1147 & 0.9170 & 0.0289 & 0.1305 & 0.9065 \\
SimVP & 0.0398 & 0.1541 & 0.8937 & 0.0347 & 0.1447 & 0.8942 & 0.0695 & 0.2027 & 0.8540 & 0.0827 & 0.2206 & 0.8443 \\
PredRNN-v2 & 0.0712 & 0.1984 & 0.8421 & 0.0958 & 0.2316 & 0.8035 & 0.1219 & 0.2638 & 0.7723 & 0.1482 & 0.2897 & 0.7498 \\
PastNet & 0.1200 & 0.2661 & 0.7637 & 0.1372 & 0.2828 & 0.7288 & 0.1497 & 0.2942 & 0.7282 & 0.1859 & 0.3240 & 0.7086 \\
Convlstm & 0.3001 & 0.4005 & 0.5044 & 0.2568 & 0.3709 & 0.5727 & 0.2462 & 0.3636 & 0.5904 & 0.2370 & 0.3568 & 0.6605 \\
\bottomrule
\end{tabular}
\end{table*}

\begin{figure*}[t] 
  \centering
  \vspace{-15pt} 
  \includegraphics[width=0.98\linewidth,keepaspectratio]{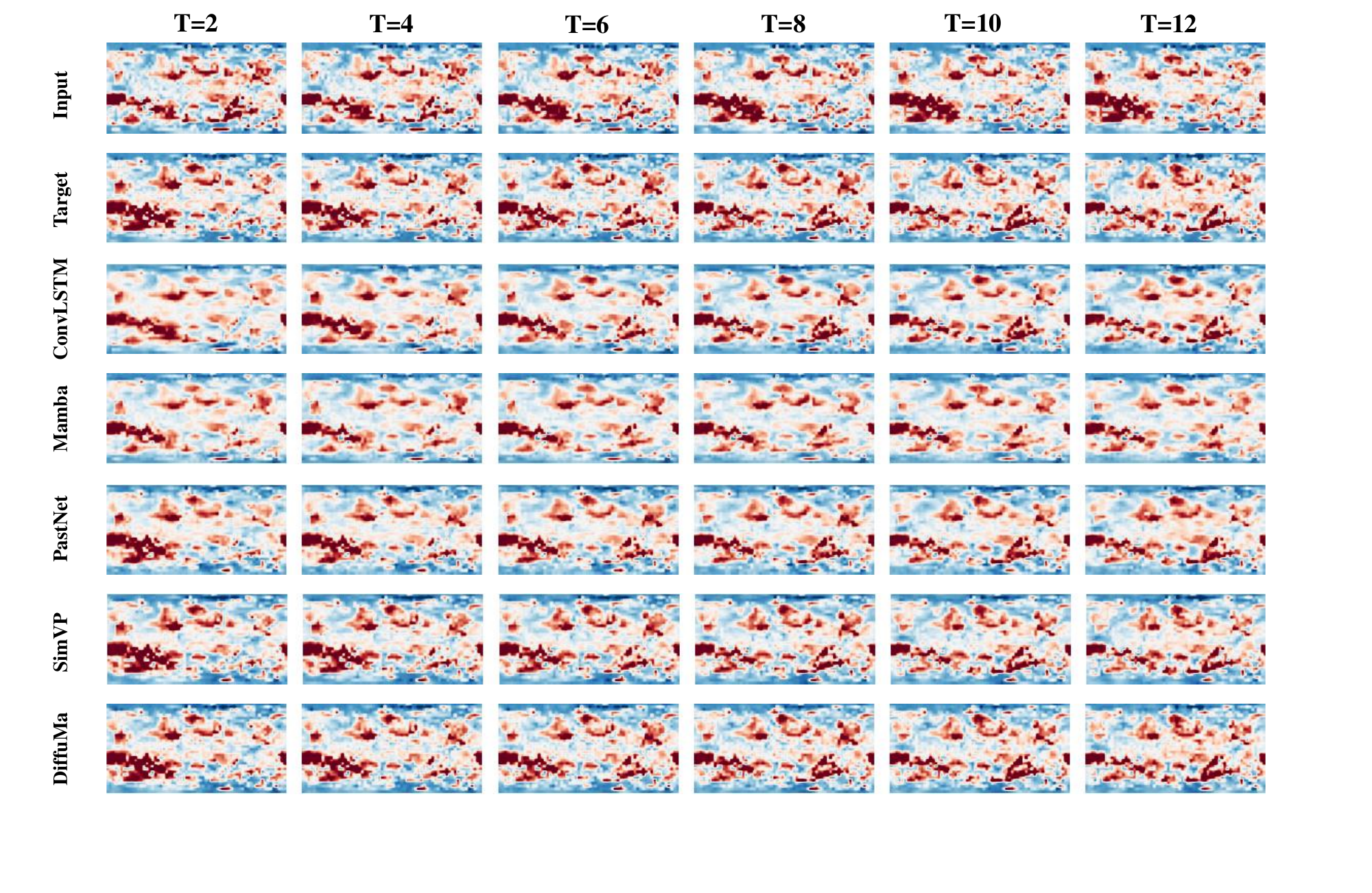}
  \caption{Comparative Analysis of Prediction Results on WeatherBench}
  \label{fig:experiment_weatherbench_comparison}
  \vspace{5pt} 
\end{figure*}
\begin{figure}[!h] 
  \centering
  \includegraphics[width=0.98\linewidth,keepaspectratio]{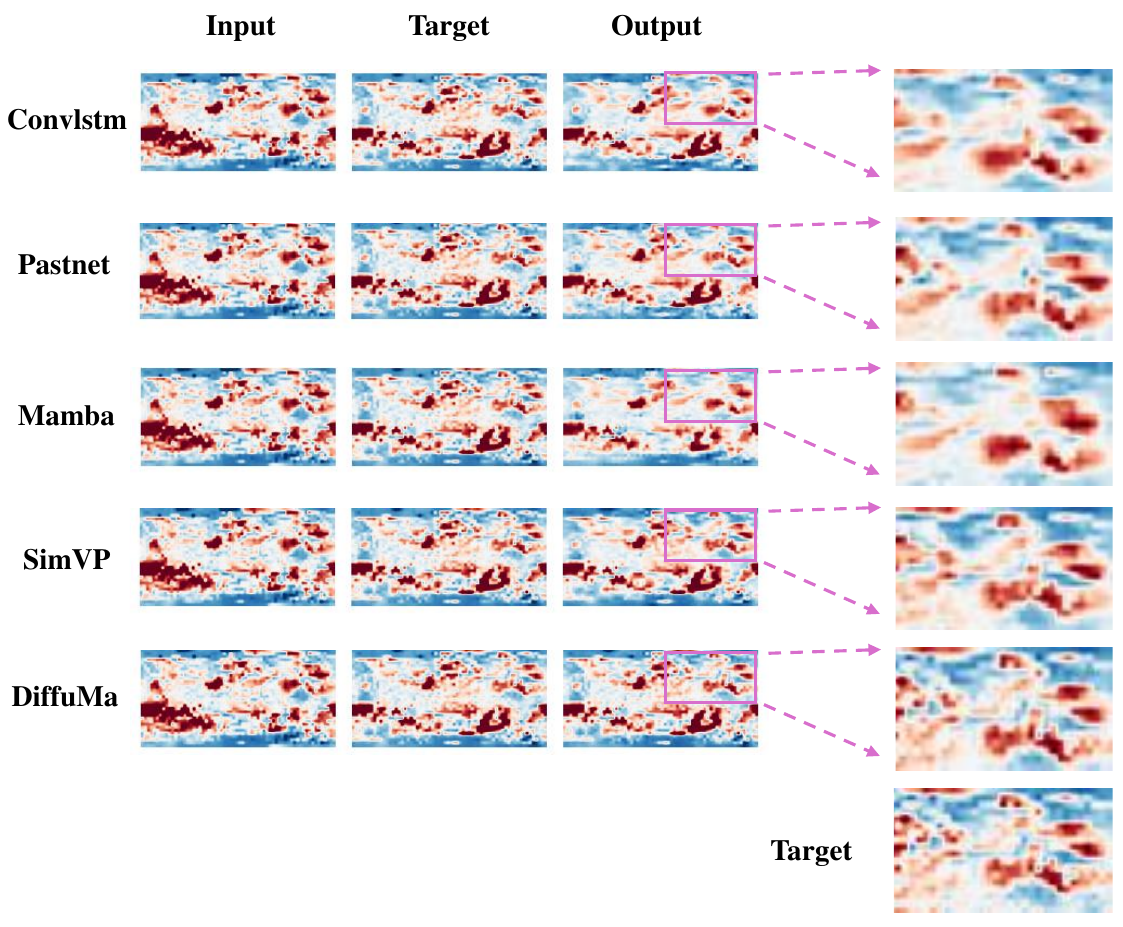}
  \caption{Comparative Analysis of Prediction Details on WeatherBench}
  \label{fig:experiment_weatherbench_detail}
  \vspace{5pt} 
\end{figure}

\subsection{Comparative Analysis of Results across Multiple Aspects}
Although DiffuMa demonstrates superior performance in both quantitative metrics and visual prediction results, comparative experiments across more dimensions remain necessary. This section presents a comprehensive comparison between DiffuMa and baseline methods across various aspects.

\subsubsection{Epoch-Varying Evaluations}
As illustrated in Table~\ref{tab:Epochs_mae_mse} and Table~\ref{tab:ssim_full}, DiffuMa maintains consistently superior predictive performance throughout the entire training process. Whether under short-term training (10 epochs) or medium-term training (100 epochs), DiffuMa significantly outperforms all baseline models across key evaluation metrics, including MSE, MAE and SSIM. This robust evidence demonstrates that DiffuMa possesses exceptional training stability and strong generalization capabilities, with its advantages being consistently validated across diverse training phases and multi-dimensional evaluation frameworks.

\subsubsection{Prediction-Length-Varying Evaluations}
As evidenced in Table~\ref{tab:PL_mae_mse_ssim}, DiffuMa demonstrates comprehensive performance dominance across full-scale prediction tasks. Specifically, it significantly surpasses all leading baseline models under four critical scenarios: short-term prediction (3-step), medium-term prediction (6-step), medium-to-long-term prediction (9-step), and long-term prediction (12-step). This consistent superiority across varying temporal scales substantiates the model's breakthrough advancements in temporal extrapolation capability and long-range dependency modeling.
\begin{figure*}[!h] 
  \centering
  \includegraphics[width=0.98\linewidth,keepaspectratio]{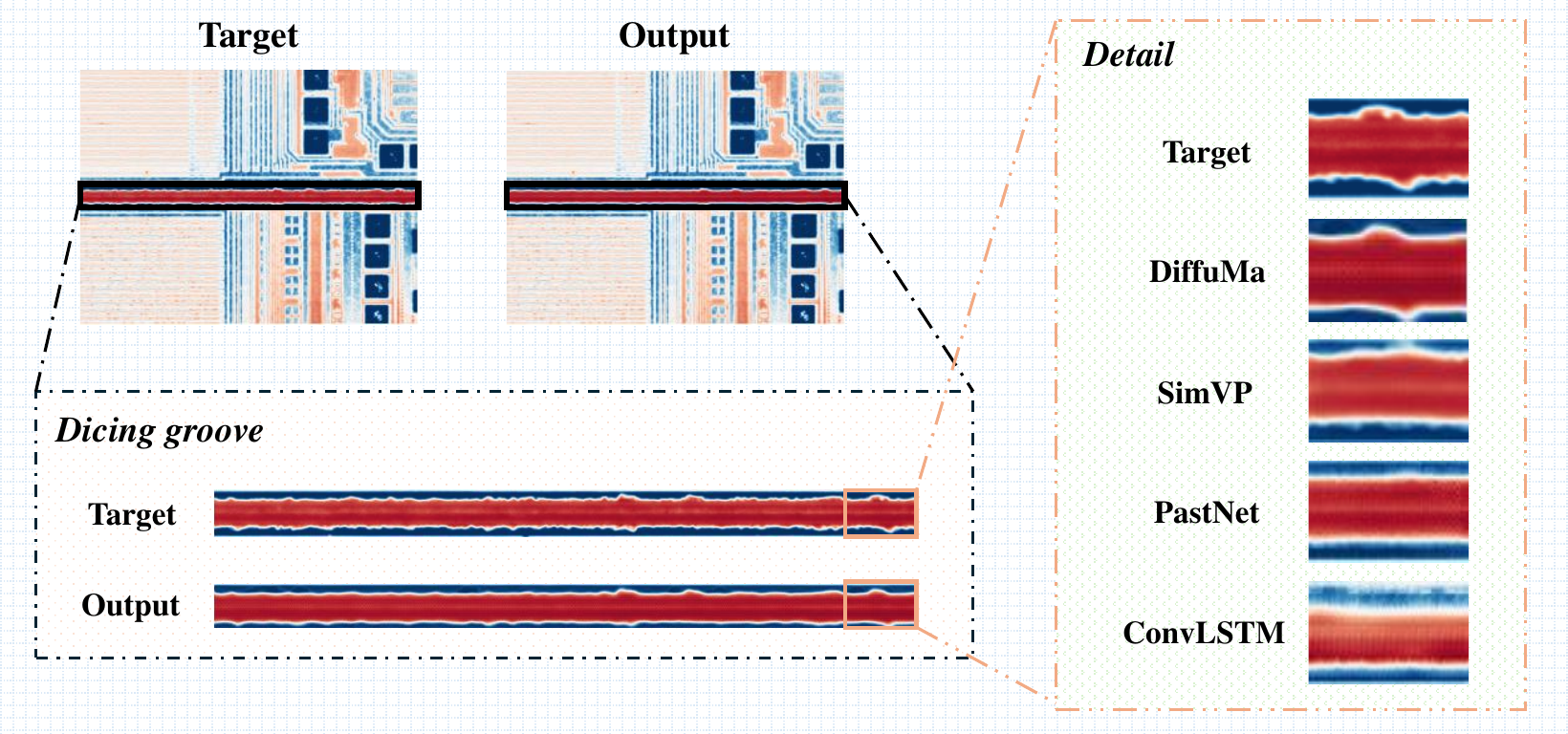}
  \caption{Comparative Analysis of Prediction Details on CHDL}
  \label{fig:experiment_CHDL_comparison}
  \vspace{5pt} 
\end{figure*}

\subsection{Visual Benchmarking of Critical Generation Details}
As demonstrated in Figure~\ref{fig:experiment_weatherbench_comparison}, through comparative experiments with other mainstream baseline models (including ConvLSTM, PastNet, SimVP, and Mamba), it is observed that DiffuMa significantly outperforms other models in terms of detail restoration capability in prediction results. Further visual evidence in Figure~\ref{fig:experiment_weatherbench_detail} and Figure~\ref{fig:experiment_CHDL_comparison}, the predicted images generated by DiffuMa exhibit superior performance in both spatial details (e.g., texture and edge sharpness) and temporal continuity (e.g., smoothness of dynamic changes). DiffuMa is capable of more accurately capturing subtle movements and morphological changes. This advantage is primarily attributed to the integration of the fine-grained generation capability of diffusion models with the efficient temporal modeling capability of the Mamba model, thereby achieving a better balance between detail preservation and computational efficiency.

\section{Conclusion}
\label{sec:conclusion}
DiffuMa achieves significant performance improvements in video prediction tasks by integrating the Mamba model with the diffusion model mechanism, while also making notable breakthroughs in computational efficiency. Experimental results demonstrate that DiffuMa not only excels on natural phenomenon datasets but also exhibits robust generalization capabilities on non-natural phenomenon datasets. Its efficient training process and superior predictive performance provide novel insights and methodologies for developing lightweight yet high-precision generative models.

Future research directions may focus on further exploring the application of DiffuMa in other complex scenarios, such as extreme weather forecasting and medical image analysis, as well as optimizing its computational efficiency to support larger-scale datasets and real-time prediction tasks.

\section*{CRediT authorship contribution statement}
\begin{itemize}
    \item Xinyu Xie: Conceptualization, Methodology, Writing -- original draft, Investigation
    \item Weifeng Cao: Writing -- review \& editing, Project administration, Funding acquisition
    \item Jun Shi: Data curation
    \item Yangyang Hu: Writing -- review \& editing
    \item Hui Liang: Writing -- review \& editing, Funding acquisition
    \item Wanyong Liang: Writing -- review \& editing, Supervision
    \item Xiaoliang Qian: Funding acquisition
\end{itemize}

\section*{Declaration of competing interest}
The authors declare that they have no known competing financial interests or personal relationships that could have appeared to influence the work reported in this paper.

\section*{Acknowledgments}
This work is supported by the National Natural Science Foundation of China (62076223);the Graduate Education Reform Project of Henan Province (2023SJGLX037Y);the Research Project of Humanities and Social Sciences of the Ministry of Education, China (No. 24YJAZH075); the Research Project of Humanities and Social Sciences of Henan Province, China (No. 2025-ZZJH-370).

\section*{Data availability}
Data will be made available on request.
\bibliographystyle{elsarticle-num-names} 
\bibliography{refs}

\end{document}